\documentclass[runningheads]{llncs}
\usepackage[T1]{fontenc}
\usepackage{graphicx}
\usepackage{xcolor}

\begin{document}
%
\title{Unsupervised Domain Adaptation for Medical Image Segmentation via Feature-space Density Matching}
\titlerunning{UDA for Semantic Segmentation via Feature-Space Density Matching}
%
\author{Tushar Kataria\inst{1,2} \and Beatrice Knudsen\inst{3} \and
Shireen Elhabian \inst{1,2} }

\authorrunning{T. Kataria et al.}
%
\institute{Kahlert School of Computing, University Of Utah \and
Scientific Computing and Imaging Institute, University of Utah \and Department of Pathology, University of Utah \\ \{tushar.kataria,shireen\}@sci.utah.edu, beatrice.knudsen@path.utah.edu }
\maketitle              
\begin{abstract} 
Semantic segmentation is a critical step in automated image interpretation and analysis where pixels are classified into one or more predefined semantically meaningful classes. Deep learning approaches for semantic segmentation rely on harnessing the power of annotated images to learn features indicative of these semantic classes. Nonetheless, they often fail to generalize when there is a significant domain (i.e., distributional) shift between the training (i.e., \textit{source}) data and the dataset(s) encountered when deployed (i.e., \textit{target}), necessitating manual annotations for the target data to achieve acceptable performance. This is especially important in medical imaging because different image modalities have significant intra- and inter-site variations due to protocol and vendor variability.
Current techniques are sensitive to hyperparameter tuning and target dataset size. This paper presents an unsupervised domain adaptation approach for semantic segmentation that alleviates the need for annotating target data. Using kernel density estimation, we match the target data distribution to the source in the feature space, particularly when the number of target samples is limited (3\% of the target dataset size). We demonstrate the efficacy of our proposed approach on 2 datasets, multisite prostate MRI and histopathology images. 
\end{abstract}

\section{Introduction} 
Human visual systems classify and delineate (segment) every object present in their environment. Segmentation is especially important in medical imaging because of the highly specific domain knowledge required to outline the relevant objects (e.g., tumor, disease tissue, cancer) \cite{liu2023deep}. 
Accurately identifying the exact boundaries of these objects (or the size of the tumor) is necessary for reliable and interpretable automation of disease diagnosis, analysis, and treatment planning  \cite{wang2012clinical}. 

When trained with a representative and sufficient quantity of training data, deep learning models consistently make more accurate predictions. 
However, these models can focus on learning spurious signals \cite{degrave2021ai} rather than features of actual disease pathology. Deep learning models learn low-level texture features more than high-level shape/morphological features \cite{hermann2020origins}, which impacts the performance of the learned model (trained on \textit{source dataset}) when new data with different low-level data statistics (\textit{target dataset}) are introduced during inference. This is called a distributional (or domain) shift in the input dataset. Such a shift results in a loss of precision and trust in the model's predictions based on the new data. Even minor distributional shifts where input images are sketches of the same objects have shown significant drops in performance \cite{wang2019learning}. This domain shift is problematic in medical imaging \cite{liu2020ms} because access to large amounts of training data is limited and hence we have to rely on pre-trained models trained.


Domain adaptation methods have been suggested as a solution to address the performance decline of models caused by domain shifts \cite{tzeng2017adversarial,vu2019advent,liu2022deep}. Various techniques have been proposed for supervised and unsupervised domain adaptation (UDA) depending on the availability of annotations in the target domain \cite{toldo2020unsupervised,wang2020alleviating,liu2022deep}. UDA techniques is more advantageous since pixel-wise annotations for segmentation tasks, particularly in the context of medical images, are prohibitively expensive due to the specialized knowledge required \cite{aljabri2022towards}.

UDA approaches for semantic segmentation can be broadly categorized into three classes. First is \textit{adversarial domain adaptation} \cite{tzeng2017adversarial,bolte2019unsupervised,haq2020adversarial}, which aims to learn domain-independent backbone features by maximizing the domain classification loss using source and target features and passing a negative gradient to the feature extraction backbone via a gradient reverse layer. Second is \textit{Fourier domain adaptation} \cite{yang2020fda,xu2021fourier}, which uses Fourier domain transformations to adapt frequency amplitude based on the assumption that phase information between domains does not change.
Third is \textit{density matching}, where the source and target densities of either input space \cite{bousmalis2017unsupervised}, output space \cite{tsai2018learning}, or feature space \cite{long2015learning} are matched. \cite{bousmalis2017unsupervised} used conditional GANs (generative adversarial networks) to transform images of source dataset to look like target dataset, whereas \cite{tsai2018learning} and \cite{long2015learning} only use discriminator for density matching between source and target features. Density matching with other penalties, such as maximum mean discrepancy (MMD) \cite{li2017mmd,erkent2020semantic,kumagai2019unsupervised} or Wasserstein GAN \cite{erkent2020semantic}, has also been tried. Adverserial-based approaches are highly sensitive to hyperparameter selection \cite{tzeng2017adversarial,bolte2019unsupervised}. 
Fourier domain adaptation frameworks are sensitive to frequency space selection and mixing ratio. 
Density-matching approaches are highly sensitive to hyperparameters \cite{li2017mmd,erkent2020semantic}, and are difficult to train because of minimax games. They also require large amounts of target data to converge. 

Here, we present a novel unsupervised domain adaptation for medical image segmentation, leveraging (1) nonparametric density estimation via kernel density estimation (KDE) and (2) matching density via Jenson-Shannon divergence (JSD), for adapting learned features in segmentation networks. KDE \cite{wkeglarczyk2018kernel,kim2012robust} has been shown to perform better for generative modeling for smaller datasets \cite{saha2022gens}. Hence, KDE offers a more stable solution for matching source and target feature densities, compared to adversarial learning, especially in low-sample size scenarios, which is typical in medical imaging. The nonparametric nature of KDE provides a rich training signal for domain adaptation compared with MMD \cite{li2017mmd}, which uses only moments to match density. Furthermore, KDE allows for batch-wise density matching during training, which matches the full density in the feature space through the batched samples. The kernel bandwidth is estimated by randomly drawing training samples and mapping them to the feature space. The estimated densities of the source and target datasets are matched using JSD loss. The proposed method regularizes the feature space resulting in the model learning generic features that are domain independent.

We compare our results with other density matching methods such as MMD \cite{li2017mmd,kumagai2019unsupervised}, using constant bandwidth as done in other proposed techniques  \cite{li2017mmd,erkent2020semantic}.
We also compare our results with adversarial training \cite{tzeng2017adversarial,bolte2019unsupervised} and density matching using discriminator in feature space \cite{long2015learning} as well as output space \cite{tsai2018learning}. Our method is closely related to feature space \cite{long2015learning} and output space \cite{tsai2018learning} density matching but instead of using a discriminator for density matching, we use JSD for divergence and KDE for estimating the underlying feature probability distribution. We follow the methods listed in the respective papers to implement our own versions for comparison. 
The contributions of this paper are as follows:-
\begin{itemize}
    \item We propose a novel approach for unsupervised domain adaptation for semantic segmentation that is based on a rich (nonparametric) representation of the underlying feature distribution. 
    \item We demonstrate that our proposed methods statistically outperform other methods for density matching with small target dataset sizes (3\% or 30\% of target dataset size). 
    \item We demonstrate the efficacy of the proposed approach on different modalities (histopathology \cite{graham2019mild,sirinukunwattana2017gland} and multi-site MRI \cite{liu2020ms}), supported by ablation experiments to assess the impact of feature space choice, frequency of bandwidth estimation, target data sample size, and the number of KDE samples.
\end{itemize}
\section{Methodology}\label{section:methodology}

\subsection{Problem Setup}

\begin{figure}[!htb]
    \centering
     \includegraphics[scale=0.42]{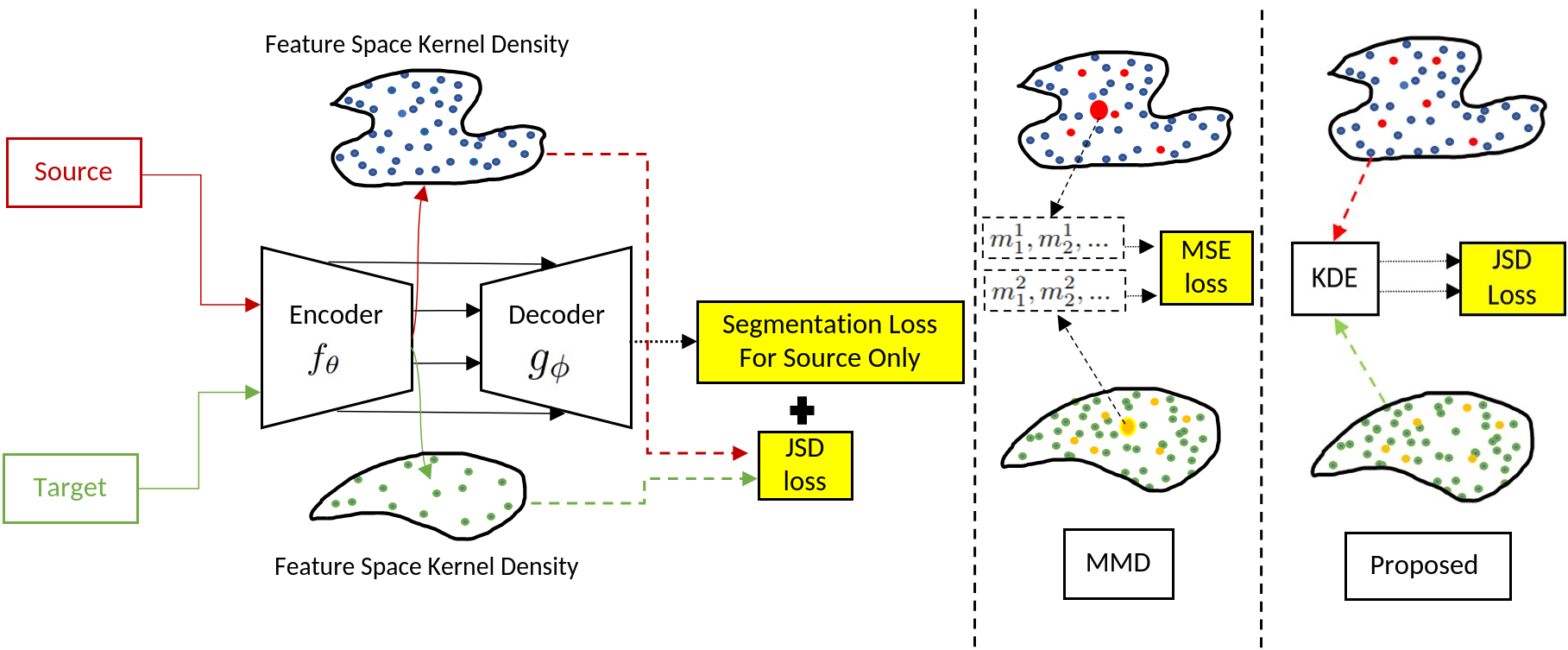}
    \caption{\textbf{Block diagram of the proposed methodology.} The model is assumed to be a standard encoder-decoder configuration with skip connections and trained using annotations only from the source dataset. Deep features are extracted from both the source and the target to estimate their densities using KDE, whereas MMD only matches moments while the proposed method matches feature space densities estimated with KDE using a JSD loss. }
    \label{fig:my_label:1}
    \vspace{-0.2in}
\end{figure}

Most deep learning architectures for semantic segmentation follow an encoder-decoder configuration as depicted in Figure \ref{fig:my_label:1}. Let $f_{\theta}(.)$ be the encoder and $g_{\phi}(.)$ be the decoder. For an input image $\mathbf{I}$, the model performs the following operations,
$\mathbf{x} = f_{\theta}(\mathbf{I})$ and $ \mathbf{y} = g_{\phi}(\mathbf{x})$, where $\mathbf{x}$ is the encoded features in the learned feature space. For segmentation networks, we can have multiple deep feature encoding and decoding spaces, but for the sake of simplicity, we assume the deepest feature space as $\mathbf{x}$ (one with the lowest spatial resolution and highest channel resolution).
Deep learning models fail to generalize when there is a domain shift in the input space. We hypothesize that this domain shift causes a density shift in the feature space of the learned model, causing it to fail for unseen data. We propose that if the model is regularized by a density-matching loss between feature space distributions of the two domains, the model will not suffer from the same domain shift on seeing the new domain. There are two main aspects to address the feature space density matching: (1) the representation of density and (2) the density matching loss.

\textbf{Representation of density.} Density in feature space can be represented by moments (mean, variance) which assumes factorized Gaussian as the default distribution of the feature space. However, factorized Gaussian implies a limiting assumption of a unimodal, disentangled distribution in the feature space. We can also assume parametric densities following certain characteristics of multivariate Gaussian or a mixture of Gaussians. However, both of these make strong assumptions about the distribution of sample points and are not driven by data. Nonparametric methods such as KDE, on the other hand, do not make such strong assumptions and are more suited to be learned from data. Hence, these methods can provide a richer and more flexible description of the feature space density. 

\textbf{Density matching loss.} KL divergence is asymmetric property so may not be suitable for domain adaptation applications because it's a uni-directional loss. JSD, on the other hand, is symmetric, which tries to regularize source features to stay close to the target and vice-versa.

 
 \subsection{Unsupervised Domain Adaptation via KDE}

The block diagram of our proposed methodology is shown in Figure \ref{fig:my_label:1}. The segmentation model is trained using annotations from only the source dataset. In our setting, no annotations from the target dataset are used. 
Density matching loss acts as a regularizer, making sure that the feature distribution of the source and target datasets does not diverge from each other.  The network is thus trained with a loss given by
\begin{equation}
    \mathcal{L} = \mathcal{L}_{(seg, source)} + \lambda JSD[p_s,p_t]
\end{equation}
where $\mathcal{L}_{(seg, source)}$ is the supervised segmentation loss on the source dataset, and $\lambda$ is a hyperparameter that defines the contribution of the density matching loss, and $p_s$ and $p_t$ are density estimates for source and target dataset, respectively.

\textbf{Kernal density estimation.} Let $\mathbf{x}_1, \mathbf{x}_2, \mathbf{x}_3, ..., \mathbf{x}_N$ be the number of sampled points from the encoded feature space. The kernel density estimate $p_{est}(\mathbf{x})$ can be written as : 
\begin{equation}\label{eqn:kde}
p_{{est}}(\mathbf{x}) = \frac{1}{N} \sum_{n=1}^{N} K\left(\frac{\|\mathbf{x}-\mathbf{x}_{n}\|_2}{\sigma}\right)
\end{equation}
where $K$ is assumed to be a Gaussian kernel in our experiments. The bandwidth parameter ($\sigma$) is estimated to be the mean of the distance between the nearest neighbors in the feature space. 
\begin{equation}
    \sigma = \frac{1}{N} \sum_{n=1}^{N} \| \mathbf{x_n}-\gamma({\mathbf{x}_n})\|_2^2 
\end{equation}
where $\gamma({\mathbf{x}_n})$ returns the nearest neighbor of $\mathbf{x}_n$. As bandwidth is estimated from the data, the method used for estimating feature distribution remains non-parametric. Using Eq. \ref{eqn:kde}, we estimate the density of the source ($p_{s}$) and target ($p_{t}$) datasets using the same kernel but with different bandwidth parameters obtained from their respective feature spaces. JSD loss is calculated using
\begin{equation}
{JSD}[p_{s},p_{t}] = \frac{1}{2}\big\{{{KL}}[p_s,M] + {{KL}}[p_t,M]\big\}, M = \frac{p_s+p_t}{2}
\end{equation}
where $KL$ is the KL-divergence between the two distributions. 

\section{Results and Discussion}
\subsection{Experimental Setup}
\textbf{{Datasets.}} We used datasets for gland segmentation in histopathology images and prostate segmentation in a multisite MRI dataset. Two datasets CRAG \cite{graham2019mild} and GlaS \cite{sirinukunwattana2017gland} are used for gland segmentation in the colon histology dataset. 
A multisite MRI dataset \cite{liu2020ms} from six different sites, with different field strengths (3 and 1.5 Tesla) and different vendors, was used with different source and target configurations. This enabled us to test multisource, multitarget, as well as held-out target settings.


\noindent \textbf{Training setup and hyperparameters.} Networks are trained for 5 different train/validation data splits and respective performance(using dice scores \cite{sirinukunwattana2017gland}) mean and standard deviations are reported when trained from scratch with Gaussian initialization which factors in the stochasticity that may be caused due to training and validation data used for source training and density estimations.

For density estimation, the number of samples for KDE is set to 20. KDE points are sampled every 5 epochs for bandwidth estimation. $\lambda$ is set to 0.01 for the histopathology dataset and 0.001 for the MRI dataset based on performance on the validation set. 
For all experiments presented here the learning rate, weight decay, batch size, and number of epochs are set to 1e-4, 1e-4, 10, and 1000, respectively. All models were trained using the PyTorch framework on Nvidia A30(24 GB) GPUs. The optimal hyperparameters for the segmentation model were selected based on the performance of the validation set. 20 \% of the training set was fixed as the validation set. The best model is saved with the lowest validation loss. The test set is used only at the end of the training process after hyperparameter optimization. We will release the associated code and models at a later date for open-source usage.

\noindent \textbf{Latent space for density estimation}
 We used a U-Net with skip connections with up to 5 decompositions for our experiments. The weights were initialized using Pytorch default initialization. Density matching is performed at the deepest encoding layer which latent dimension of 1024x8x8, which is unrolled to a 65536x1 vector. Ablation experiments with different choices of latent spaces are also reported in the Supplementary Table \ref{tab:dice_score:FeatureSpace}. 

\subsection{Results}
\begin{figure}[!htb]
    \centering
    \includegraphics[scale=0.41]{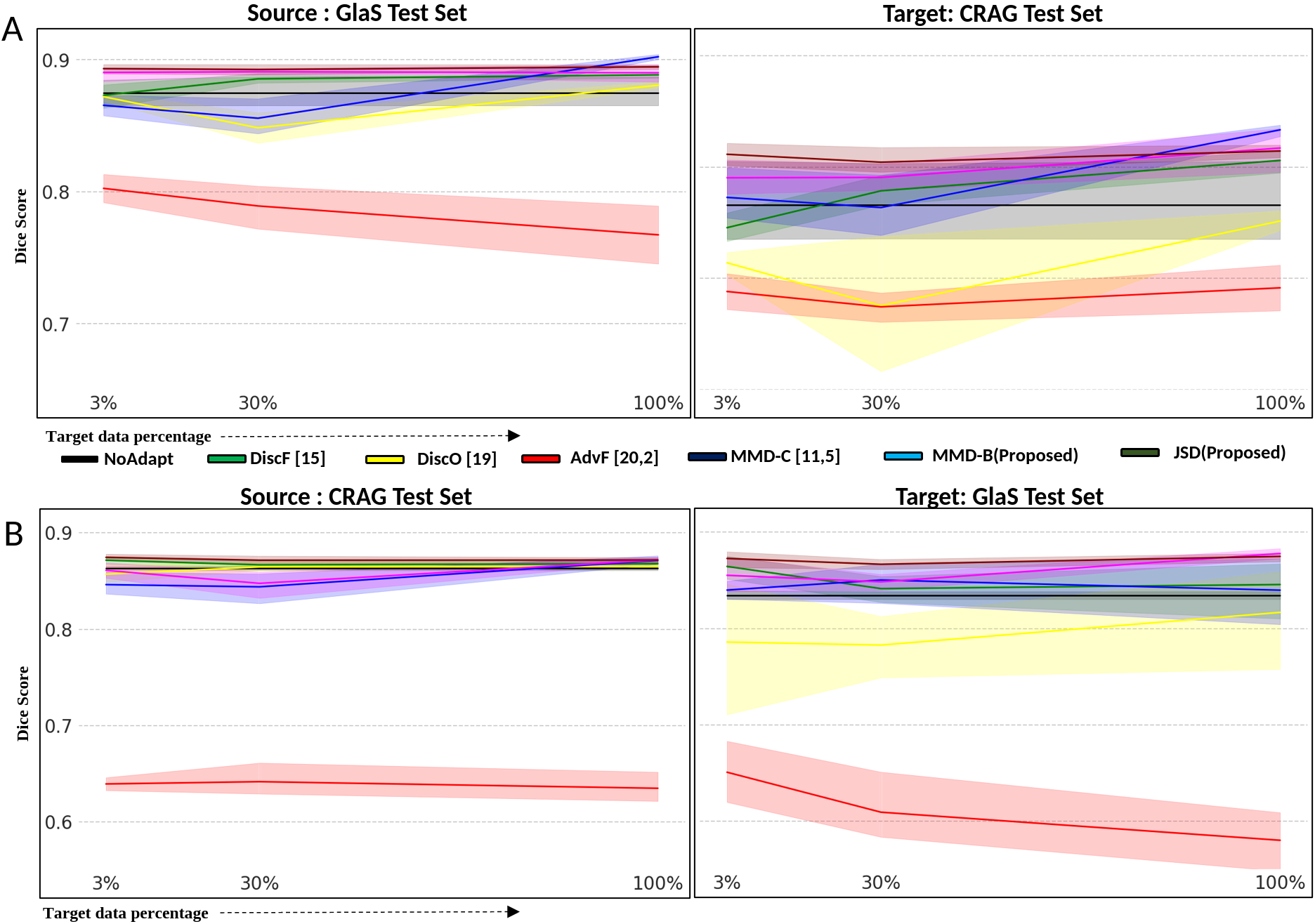}
    \caption{\textbf{UDA results on gland segmentation.} A) GlaS dataset is the source and CRAG is the target. B) CRAG dataset is the source and GlaS as the target. Samples from the training set are used for domain adaptation, the test set is only used for evaluation of the trained models.  Method proposed outperforms other comparisons at low target dataset size with both 3\% and 30\% of the data. \textbf{No Adapt} refers to the generic case where a model is trained on the source dataset and tested on the target dataset without any finetuning. We treat this case as a baseline because we are targeting UDA, where we cannot access target dataset annotations. For other baselines, 
here \textbf{Adver} refers to \cite{tzeng2017adversarial,bolte2019unsupervised}, \textbf{DiscF} refers to discriminator in feature space \cite{long2015learning}, and \textbf{DiscO} refers to discriminator in output space \cite{tsai2018learning}.  \textbf{MMD-C} using constant bandwidth as done in previous proposed techniques \cite{li2017mmd,erkent2020semantic}. \textbf{MMD-B} using bandwidth proposed in Section 2.2 above and, \textbf{JSD} is the proposed method. We can clearly observe that average performance on target data with the proposed method is higher than other comparisons for low target dataset setting and competitive when using all the data.}
\label{fig:my_label:DomainAdaptationResults}
\vspace{-0.25in}
\end{figure}
\textbf{Gland segmentation results.} 
Figure \ref{fig:my_label:DomainAdaptationResults} shows significant drop in performance without any domain adaptation. Our proposed method does not have any effect on the accuracy of the source model for any target dataset size, whereas in other methods we observe variation in performance on the source dataset as a function of the target dataset size. In addition, our proposed method achieves higher accuracy on the source dataset than other methods. 

\begin{figure}[!htb]
    \centering
    \includegraphics[scale=0.32]{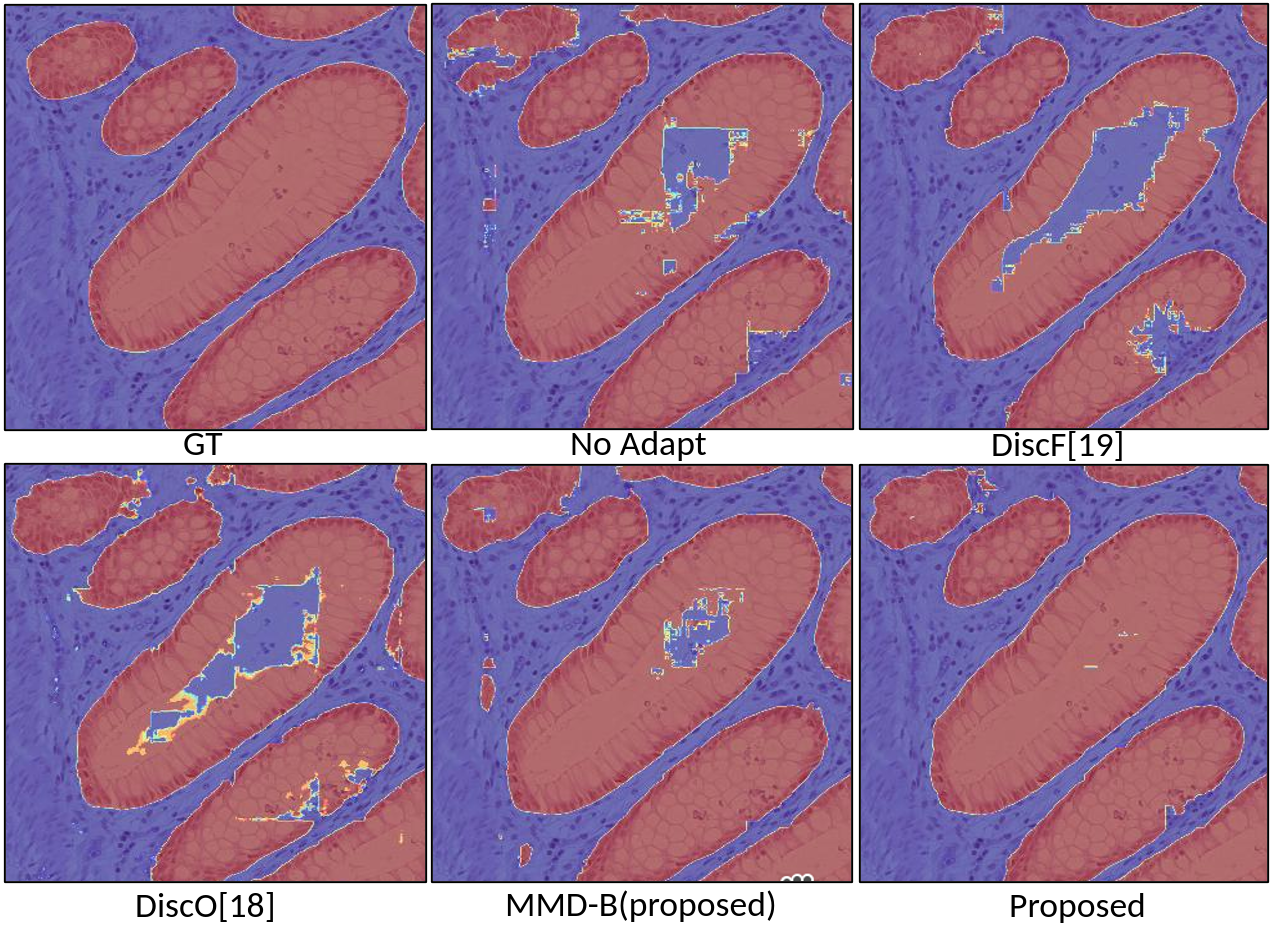}
    \vspace{-0.1in}
    \caption{\textbf{Qualitative results for gland segmentation.} Results for the target dataset (CRAG) when the model is trained using the GlaS dataset as the source. The proposed method gives better quality results compared to other techniques when using a small amount of the target dataset. }
    \label{fig:my_label:qualitative_results_glands}
\end{figure}

On target datasets, our proposed method outperforms all other comparisons when using only 3\% or 30\% of the target dataset, with MMD with the proposed bandwidth being the closest second. Only MMD with a constant bandwidth(MMD-C) outperforms our proposed method when using a 100\% target dataset when the source dataset is GlaS and the target dataset is CRAG. In all other cases, the proposed method outperforms others. The performance gain is higher when the source dataset is CRAG compared to GlaS, which can be attributed to the difference in the number of samples. The CRAG dataset has more samples, which can result in the model getting more biased toward the source dataset. However, the proposed method successfully helps overcome that bias, resulting in a higher gain. Qualitative results are shown in Figure \ref{fig:my_label:qualitative_results_glands}. Other methods struggle to segment the correct outline of the glands, confusing pixels inside the glands with the background. However, the proposed methods correctly segment all pixels inside glands as the correct class. All Ablation studies were performed on these datasets (results are reported below).

\textbf{Multisite prostate segmentation results.} For the multisite MRI data, prostate segmentation data is available for 6 sites. Hence, we modified the testing methodology to have an out-of-distribution (held-out) domain that is not shown to any network during training or UDA. This setup helps in gauging whether the proposed UDA methodology can improve the model's prediction for an unseen dataset. We combine these 6 datasets in multiple sources, targets, and held-out datasets sets. Table \ref{tab:dice_score:MRI1}(\ref{tab:dice_score:MRI3},\ref{tab:dice_score:MRI2} and \ref{tab:dice_score:MRI4} in supplementary) show results for different configurations mentioned above. Table \ref{tab:dice_score:MRI1} represents single source multitarget configuration. We can observe that not only do our proposed methods (MMD-B and JSD) outperform on the source and multitarget datasets, but also on the held-out dataset. Similar results are observed for multisource single target configuration shown in Table \ref{tab:dice_score:MRI3} in Supplementary. Our proposed approach not only maintains the accuracy in source and target dataset but also has consistently higher accuracy on held-out datasets. Qualitative results are shown in supplementary Figure \ref{fig:MRI:Results}. 
\begin{table}[!htb]
\small
\caption{\textbf{Single source multitarget configuration:} Mean of Dice Scores on source and target datasets. When only 3\% of the target dataset is used for distribution estimation. We see that not only do our proposed methods (MMD-B and JSD) outperform on the source and multiple target datasets, but also on the held-out dataset.}
\centering
\setlength{\tabcolsep}{4pt}
\begin{tabular}{ l|c|cccc|c } 
 & \bf Source & \multicolumn{4}{c|}{\bf Target } & \bf Held Out  \\
 \hline
 & \bf  \textbf{RUNMC} & \bf \textbf{BMC} & \bf \textbf{I2CVB}  & \bf \textbf{UCL} & \bf BIDMC & \bf HK   \\
\hline
No Adapt  & 0.87 &	 0.79	&  0.63 &	 0.74 &  0.53 &  0.649\\ 
\hline
Adver  \cite{bolte2019unsupervised}& 0.717 &	 0.623	&  0.61 &	 0.564 &  0.488 &  0.496\\ 
DiscF \cite{long2015learning}   & 0.902 &   0.795	&  0.622 &	 0.812 &  0.547 &  0.667\\ 
DiscO  \cite{tsai2018learning} & 0.878 &	 0.765	&  0.604 &	 0.768 &  0.518 &  0.626\\
MMD-C  & 0.917 &0.835	 &  0.63 &	0.822  & 0.568 & 0.692 \\ 
\hline
MMD-B   &  0.911 &	 0.824	&  \bf 0.646 &	  0.822 &   0.568 &   0.69\\ 
JSD  & \bf 0.918 &	 \bf 0.83	&   0.615 & \bf  0.829 &  \bf 0.58 &  \bf 0.711\\ 
\hline
\end{tabular}
\vspace{-0.2in}

\label{tab:dice_score:MRI1}
\end{table}



\noindent \textbf{Feature space ablation.}  We experimented with using different feature spaces of the segmentation model to do feature density estimation. We observed the difference between the test metrics for source and target datasets is not statistically different. Results for different dimensionality of the feature space density estimation are shown in Supplementary Table \ref{tab:dice_score:FeatureSpace}.

\noindent \textbf{Frequency of bandwidth estimation.} Changing the frequency of bandwidth estimation from 1, 5, 25, and 125 epochs does not show a significant change in target performance metrics. Optimal values are obtained for frequency epoch 5;  results are shown in Table \ref{tab:dice_score:GlasCragFrequency} in Supplementary.

\noindent \textbf{Number of KDE samples used.} Ablation with different numbers of KDE samples used for density estimation are shown in supplementary Table \ref{tab:dice_score:GlasCragKDESamples}. We can clearly observe that for the histopathology dataset, 20 KDE samples give the best results.


\section{Conclusion and Future Work}
We proposed a technique for unsupervised domain adaptation based on density matching and nonparametric density estimate. We showed the efficacy of the proposed approach  on 2 modalities, histopathology and multi-site MRI. The proposed technique not only improves results on target datasets but also showed consistent improvement in source and held-out results. Evaluating whether performing density matching in more than one feature space can help a model acquire a more accurate representation is a topic for future research. Although the proposed method is insensitive to hyperparameters, it does require an appropriate choice of the number of KDE points and kernel bandwidth for the dataset. Another direction for future work would be to make these hyperparameters inherently dependent on the feature space's diversity. 


%
%
%
%

\bibliographystyle{splncs04}
\bibliography{main}

\newpage
\section{Supplementary}
\vspace{-0.3in}
\begin{table}[!htb]
\caption{\textbf{Multisite source and single target configuration.} The mean Dice scores are reported when using only 3\% of the target dataset. For multi-source, single-target configurations our proposed methods outperform others on target and held-out datasets (especially on \textit{HK}).}
\vspace{-0.1in}
\centering
\setlength{\tabcolsep}{4pt}
\begin{tabular}{ l|ccc|c|cc } 
 & \multicolumn{3}{c|}{\bf Source } &  \bf Target  & \multicolumn{2}{c}{\bf Held Out}  \\
 \hline
 & \bf  \textbf{RUNMC} & \bf \textbf{I2CVB} & \bf \textbf{BIDMC}  & \bf \textbf{UCL} & \bf BMC & \bf HK   \\
\hline
No Adapt                             & 0.86 &	0.87	&  0.70 &	 0.79 &  0.83 &  0.70\\ 
\hline
Adver\cite{bolte2019unsupervised}    & 0.64 &	0.73	&  0.57 &	 0.545  &  0.611 &  0.516\\ 
DiscF \cite{long2015learning}        & 0.9  &	0.91	&  0.71 &	 0.82  &  0.86 &  0.69\\ 
DiscO\cite{tsai2018learning}         & 0.87 &	0.90	&  0.71 &	 0.80  &  0.84 &  0.66\\ 
MMD-C                                &  0.91    &	\bf 0.927	&  0.747 &	 0.854  &  0.876 &  0.722\\ 
\hline
MMD-B                                &  0.912 & 0.925	&  \bf 0.778 &	 \bf 0.873 &  0.876 & 0.737 \\ 
JSD                                  &  \bf 0.916 &  0.926	&  0.751 &	 0.867 &  \bf 0.88 & \bf 0.751 \\ 
\hline
\end{tabular}
\vspace{-0.3in}
\label{tab:dice_score:MRI3}
\end{table}

\begin{table}[!htb]
\caption{ \textbf{Multi-source multi-target configuration.} The mean Dice scores are reported when using only 3\% of the target dataset. For this configuration the performance for MMD with proposed bandwidth and the proposed density matching are comparable, although outperforming constant bandwidth and other methods.}
\small
\centering
\setlength{\tabcolsep}{4pt}
\begin{tabular}{ l|cc|cc|cc } 
 & \multicolumn{2}{c|}{\bf Source } & \multicolumn{2}{c|}{\bf Target } & \multicolumn{2}{c}{\bf Held Out}\\
 \hline
 & \bf  \textbf{RUNMC} & \bf \textbf{I2CVB} & \bf \textbf{BMC}  & \bf BIDMC & \bf \textbf{UCL}  & \bf HK   \\
\hline
No Adapt                           & 0.85     & 0.89      & 0.79   &  0.50 &	 0.66 &  0.51\\ 
\hline
Adver\cite{bolte2019unsupervised}  & 0.645    &	0.757     &  0.604  &	 0.49 &  0.551 &  0.496\\ 
DiscF\cite{long2015learning}       & 0.903    &	0.909     &  0.846  &	 0.549 &  0.841 &  0.601\\ 
DiscO \cite{tsai2018learning}      & 0.847    &	0.889	  &  0.763  &	 0.514 &  0.741  &   0.512\\ 
MMD-C                              &  0.909   & 0.908  &   0.85  & \bf 0.564 &	\bf 0.85  & \bf 0.625  \\
\hline
MMD-B                              & \bf 0.913 &  0.91  & \bf 0.859  &  0.561 &	0.838  &  0.597\\ 
JSD                                & \bf 0.913 & \bf 0.913 &  0.858	&  0.559 &	 0.836  &  0.607\\ 
\hline
\end{tabular}
\label{tab:dice_score:MRI2}
\vspace{-0.3in}
\end{table}

\begin{table}[!htb]
\caption{\textbf{Second multisource multitarget configuration.} For this configuration MMD with the proposed bandwidth outperforms all other methods. This result show that choosing data-dependent bandwidth (i.e., an average of nearest neighbor distance) can help in domain adaptation.}
\small
\small
\centering
\setlength{\tabcolsep}{4pt}
\begin{tabular}{ l|cc|c|ccc } 
 & \multicolumn{2}{c|}{\bf Source } & \bf Target  & \multicolumn{3}{c}{\bf Held Out } \\
 \hline
 & \bf  { HK } & \bf {UCL} & \bf {RUNMC}  & \bf {BMC} & \bf BIDMC & \bf I2CVB   \\
\hline
No Adapt &  0.87  &	 0.84 & 0.78 &	0.77	&  0.56 &  0.64 \\ 
\hline
Adver\cite{bolte2019unsupervised}&  0.749   &	 0.675 & 0.681 &	0.647	&  0.617 &  0.514 \\ 
DiscF \cite{long2015learning}  &  0.876  &     0.869 & 0.803 &	0.787	&  0.66 &  0.554 \\ 
DiscO\cite{tsai2018learning}   &  0.81   &	 0.77 & 0.74 &	0.71	&  0.53 &  0.63 \\ 
MMD-C   &  0.842 &  0.841 &  0.772 &	 0.748	&   0.65 &	  0.552  \\ 
\hline
MMD-B   &   0.887 &  \bf 0.889 & \bf 0.814 &	\bf 0.817	&  \bf 0.678 &	 \bf 0.579  \\ 
JSD    &  \bf 0.888 &   0.886 &  0.813 &	0.813	&  0.667 &	  0.561 \\ 
\hline
\end{tabular}
\label{tab:dice_score:MRI4}
\vspace{-0.2in}
\end{table}

\begin{table}[!htb]
\small
\caption{\textbf{Feature space ablation.} Mean and standard deviation of performance for density matching for different encoder and decoder feature spaces. We can note that performance measures are not statically significant for different feature spaces. }
\centering
\setlength{\tabcolsep}{2.5pt}
\begin{tabular}{ l|c|cc } 
  & \multicolumn{2}{c}{\bf CRAG as Source dataset }   \\
 \hline
\bf Feature Space & \bf Feature Dimensions &\bf \textbf{Source}  & \bf \textbf{Target}   \\
\hline
Deepest      & 1024x8x8& 0.87 $\pm$0.003 & \ 0.876 $\pm$0.012       \\
\hline
ENC1        & 128x128x128 &      0.871 $\pm$0.005 & 0.877 $\pm$0.015       \\ 
ENC2        & 256x64x64   &     0.874 $\pm$0.001 & 0.876 $\pm$0.009       \\ 
ENC3        & 512x32x32   &     0.872 $\pm$0.003 & 0.874 $\pm$0.009       \\
ENC4        & 1024x16x16  &     0.871 $\pm$0.001 & \bf 0.877 $\pm$0.004       \\
DEC4        & 1024x16x16  &     0.869 $\pm$0.014 &  0.874 $\pm$0.009       \\
DEC3        & 512x32x32   &     0.870 $\pm$0.007 &  0.876 $\pm$0.011       \\
DEC2        & 256x64x64   & \bf 0.875 $\pm$0.003 &  0.875 $\pm$0.013       \\
DEC1        & 128x128x128 &  0.866 $\pm$0.013    &  0.868 $\pm$0.016       \\
\hline
\end{tabular}
\vspace{-0.5in}

\label{tab:dice_score:FeatureSpace}
\end{table}

\begin{table}[!htb]
\small
\caption{\textbf{Bandwidth estimation frequency ablation.} Mean Dice Score and standard deviations for test metrics with different frequency/epochs for bandwidth estimation. We see that if bandwidth is estimated every 5 epochs, we get optimal results for both datasets.}
\centering
\setlength{\tabcolsep}{1.5pt}
\begin{tabular}{ c|cc|cc } 
 & \multicolumn{2}{c|}{\bf GlaS as Source } & \multicolumn{2}{c}{\bf CRAG as Source }   \\
 \hline
BW Frequency & \bf  \textbf{Source} & \bf \textbf{Target} & \bf \textbf{Source}  & \bf \textbf{Target}   \\
\hline
5 epochs     & \bf 0.895 $\pm$0.003 &  0.815 $\pm$0.012      &  0.87   $\pm$0.004   & \bf 0.879 $\pm$0.004  \\
\hline
1 epoch      &  0.894 $\pm$0.007 &  0.816 $\pm$0.017     & \bf 0.874  $\pm$0.001   &  0.877  $\pm$0.003  \\
25 epoch        &  0.894 $\pm$0.004 &  0.805 $\pm$0.023   & \bf 0.874  $\pm$0.004   & 0.875 $\pm$0.004  \\
125    & 0.894 $\pm$ 0.002    & \bf 0.821 $\pm$ 0.013             & 0.866 $\pm$ 0.003       & 0.863 $\pm$ 0.043 \\ 
\end{tabular}
\vspace{-0.5in}
\label{tab:dice_score:GlasCragFrequency}
\end{table}
\begin{table}[!htb]
\small
\caption{\textbf{Number of KDE samples ablation.} Mean Dice Score and standard deviations are reported for different numbers of KDE samples used for the estimation of feature space density. 20 KDE samples are optimal for gland segmentation domain adaptation.}
\centering
\setlength{\tabcolsep}{1.5pt}
\begin{tabular}{ c|cc|cc } 
& \multicolumn{2}{c|}{\bf GlaS as Source } & \multicolumn{2}{c}{\bf CRAG as Source }   \\
 \hline
Number of KDE Samples  & \bf  \textbf{Source} & \bf \textbf{Target} & \bf \textbf{Source}  & \bf \textbf{Target}   \\
\hline
20      & \bf 0.895 $\pm$0.003 & \bf 0.81 $\pm$0.012      &  \bf 0.87  $\pm$0.004   & \bf 0.879 $\pm$0.005  \\
\hline
10   & 0.874 $\pm$0.014 &  0.784 $\pm$0.036 & 0.865 $\pm$0.008  & 0.828 $\pm$0.05 \\
20   & 0.875 $\pm$0.018 &  0.78 $\pm$0.036  & 0.851 $\pm$0.015  & 0.843 $\pm$0.026 \\
80   & 0.868 $\pm$0.018 &  0.019 $\pm$0.057 & 0.86 $\pm$0.008  & 0.861 $\pm$0.001 \\ 
\end{tabular}
\vspace{-0.5in}
\label{tab:dice_score:GlasCragKDESamples}
\end{table}
\begin{figure}[!htb]
    \centering
    \includegraphics[width=12.0cm,height=2.2cm]{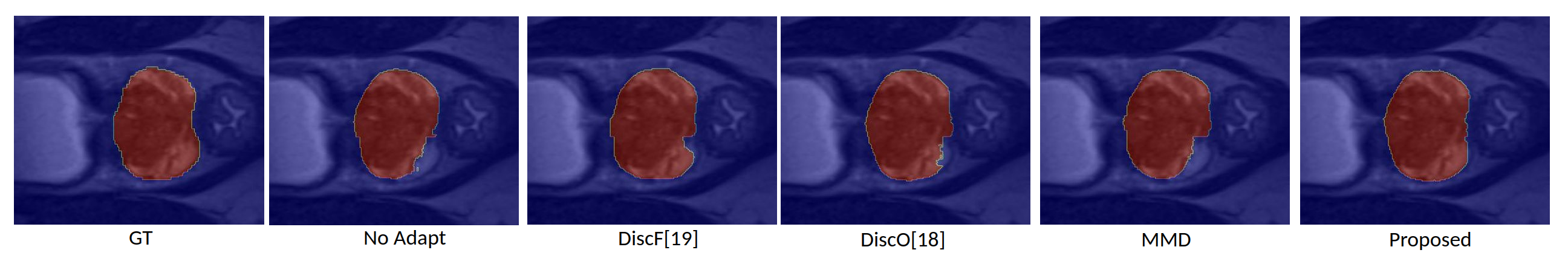}
    \caption{\textbf{Qualitative results for MRI cohort.} We obtained good quality outputs from the proposed method, which have correct outlines compared to other methods.}.
    \label{fig:MRI:Results}
    \vspace{-0.2in}
\end{figure}
\end{document}